\documentclass[12pt]{iopart}
\usepackage{iopams,bm,graphicx,cases}  
\begin{document}

\title{Traffic data reconstruction based on Markov random field modeling}

\author{Shun Kataoka$^1$, Muneki Yasuda$^2$, Cyril Furtlehner$^3$ and Kazuyuki Tanaka$^1$}

\address{$^1$Graduate School of Information Science, Tohoku University, 6-3-09 Aramaki-aza-aoba, Aobaku, Sendai, 980-8579, Japan}

\address{$^2$Graduate School of Science and Engineering, Yamagata University, 4-3-16 Jonan, Yonezawa, Yamagata, 992-8510, Japan}

\address{$^3$INRIA Saclay, LRI, B\^at. 660, Universit\'e Paris Sud, 91405, Orsay, Cedex, France}

\ead{xkataoka@smapip.is.tohoku.ac.jp}

\begin{abstract}
We consider the traffic data reconstruction problem.
Suppose we have the traffic data of an entire city that are incomplete because some road data are unobserved.
The problem is to reconstruct the unobserved parts of the data.
In this paper, we propose a new method to reconstruct incomplete traffic data collected from various traffic sensors.
Our approach is based on Markov random field modeling of road traffic.
The reconstruction is achieved by using mean-field method and a machine learning method.
We numerically verify the performance of our method using realistic simulated traffic data for the real road network of Sendai, Japan.
\end{abstract}

\maketitle

\section{Introduction}
An intelligent transportation system (ITS) is a large scale information system whose objective is to provide guidance information to drivers and to optimize transportation traffic by analyzing vehicle traffic over an entire city.
In order to provide accurate information, an ITS needs to collect accurate and comprehensive road traffic data.
Due to the development of information and sensing technologies, we can collect various types of road traffic data, density, flow, speed, and so on, from different sensing devices such as optical beacons and probe vehicles.
These sensors each have different features.
For example, a beacon, which is a fixed type of traffic sensor, can steadily collect the traffic data of the road where it is located in a shot time period; however, the detection area is narrow.
A probe vehicle, which is a GPS-equipped vehicle, can collect the traffic data of a comprehensive area, but cannot steadily collect the data and needs a long time period to acquire comprehensive traffic data.
Therefore, the fusion of various data collected from different sensors for traffic prediction has recently attracted much attention \cite{Faouzi_Leung_Kurian_InformationFusion_2011}.

Traffic prediction is a major research topic in the machine learning field \cite{Bishop_Book_2006}.
In fact, the analysis of freeway traffic has been researched since the 70s \cite{Ahmed_Cook_TransportationReserchRecord_1979}.
Travel time prediction \cite{Ide_Sugiyama_AAAI_2011}, density prediction \cite{Kriegel_Renz_Schubert_Zuefle_SIAM_2008}, and route planning \cite{Nikolova_Karger_AAAI_2008} are other active topics.
In the machine learning approach, the existence of two databases, real time (RDB) and historical (HDB), is assumed.
An RDB consists of road traffic data collected from sensors at the present time, while an HDB contains road traffic data collected from sensors and traffic surveys in the past.
The data in RDB represents a situation where we want to conduct traffic prediction while HDB consists of traffic data of roads in a comprehensive area for long time periods and can be used to help the prediction. 
That is, we use an HDB for learning and make a traffic prediction based on an RDB.

However, there still remains an important problem related to traffic prediction based on an RDB:
The quality of the prediction depends on the quality of the RDB.
We cannot acquire the complete traffic data of an entire road network in short time period since sensors are not installed on all roads.
In fact, only 22\% of the total length of trunk roads in Nagoya, Japan is covered by beacons \cite{Morikawa_Yamamoto_Miwa_Wang_Koutsuukougaku_2007}.
Further, at the present time, there are not enough probe vehicle to allow sufficient data to be acquired.
If the number of probe vehicles in Japan is a hundred thousand, we need an hour on average to acquire one or two traffic data of an entire road network \cite{Fushiki_Yokota_Kimita_Kumagai_WCITS_2004}.
Therefore, in practice, it is difficult to collect sufficiently comprehensive road traffic data in short time period to make a traffic prediction.
Therefore, we need a method to reconstruct the unobserved parts in an RDB to solve a realistic traffic prediction problem.
Recently, some researchers have tackled this problem. 
Kumagai {\it et al}. proposed a method to reconstruct the traffic data of unobserved parts in an RDB based on feature space projection \cite{Kumagai_Fushiki_Yokota_Kimita_IPSJJournal_2006}, which Kumagai and co-workers then applied to the dynamical traffic prediction problem \cite{Kumagai_Hiruta_Okude_Yokota_IPSJJournal_2012}.
In the field of statistical mechanics, Furtlehner {\it et al}. modeled road traffic as an Ising model, where the state is determined by whether a road is congested or not, and addressed the traffic reconstruction and prediction problem that arises when the observed data are incomplete using belief propagation \cite{Furtlehner_Lasgouttes_Fortelle_IEEEITS_2007}.

In this paper, we propose a new algorithm to reconstruct the traffic data of the unobserved parts in an RDB.
We use a Bayesian approach to express a posterior probability density function of unobserved roads.
Our method is based on Markov random fields (MRF) modeling of road traffic and the reconstruction of the traffic data of the unobserved parts in an RDB is achieved by solving simple simultaneous equations derived by mean-field method after learning our MRF model by utilizing HDB.
For the simplicity of the model, our method can easily address large scale problem to which we consider it difficult to apply previous methods.

The remainder of this paper is organized as follows.
In section \ref{modeling}, we introduce a graph representation of a road network and MRF modeling of road traffic.
In section \ref{reconstruction}, we propose a traffic data reconstruction algorithm based on the MRF modeling of road traffic described in section \ref{modeling}.
In section \ref{learning}, we give a framework for determining the hyperparameters in the posterior probability density function derived in section \ref{reconstruction} using the machine learning method.
In section \ref{num_exp}, we numerically verify the performances of our MRF model by using large scale simulation data for the road network of Sendai, Japan (the number of roads is $9582$).
The performances are evaluated by conducting leave-one-out cross-validation.
Finally in section \ref{disccusion}, we present our concluding remarks.

\section{MRF modeling of road traffics} \label{modeling}
In this section, we explain how road traffic is expressed by MRFs.
First, we define the undirected graph representation $(V,E)$ of a real road network. 
Let us consider a road network consisting of $N$ roads or road segments.
A vertex $i \in V := \{1, \dots, N \}$ corresponds to the $i$th road in a road network.
A set of all edges $E$ includes either edge $(i,j)$ or edge $(j,i)$ if a vehicle on road $i$ can move to road $j$ without passing along the other roads.

We assign a random continuous variable $x_i \in (-\infty, \infty)$ associated with the traffic data of road $i \in V$.
For each vertice and edge, we assign a potential function $\psi_i ( x_i )$ and $\psi_{ij} ( x_i, x_j )$, respectively.
Then, the joint probability density function of $\bm{x} := \{ x_i | i \in V \}$ is written as a product of a potential function:
\begin{eqnarray}
  P( \bm{x} ) := \frac{1}{Z} \prod_{i \in V} \psi_i ( x_i ) \prod_{(i,j) \in E} \psi_{ij} ( x_i, x_j ). \label{prob_model}
\end{eqnarray}
The quantity $Z$ is a partition function defined as
\begin{eqnarray}
  Z := \int d \bm{x} \prod_{i \in V} \psi_i ( x_i ) \prod_{(i,j) \in E} \psi_{ij} ( x_i, x_j ) \label{normalization}
\end{eqnarray}
where $\int d \bm{x}$ is taken over all the configurations of random various $\bm{x}$. 
If we want to use a discrete random variable, the integration over continuous variables in equation (\ref{normalization}) becomes a summation over discrete variables.

\begin{figure}
  \begin{center}
    \begin{tabular}{ c c }
      (a) & (b)\\
      \includegraphics[width=4cm]{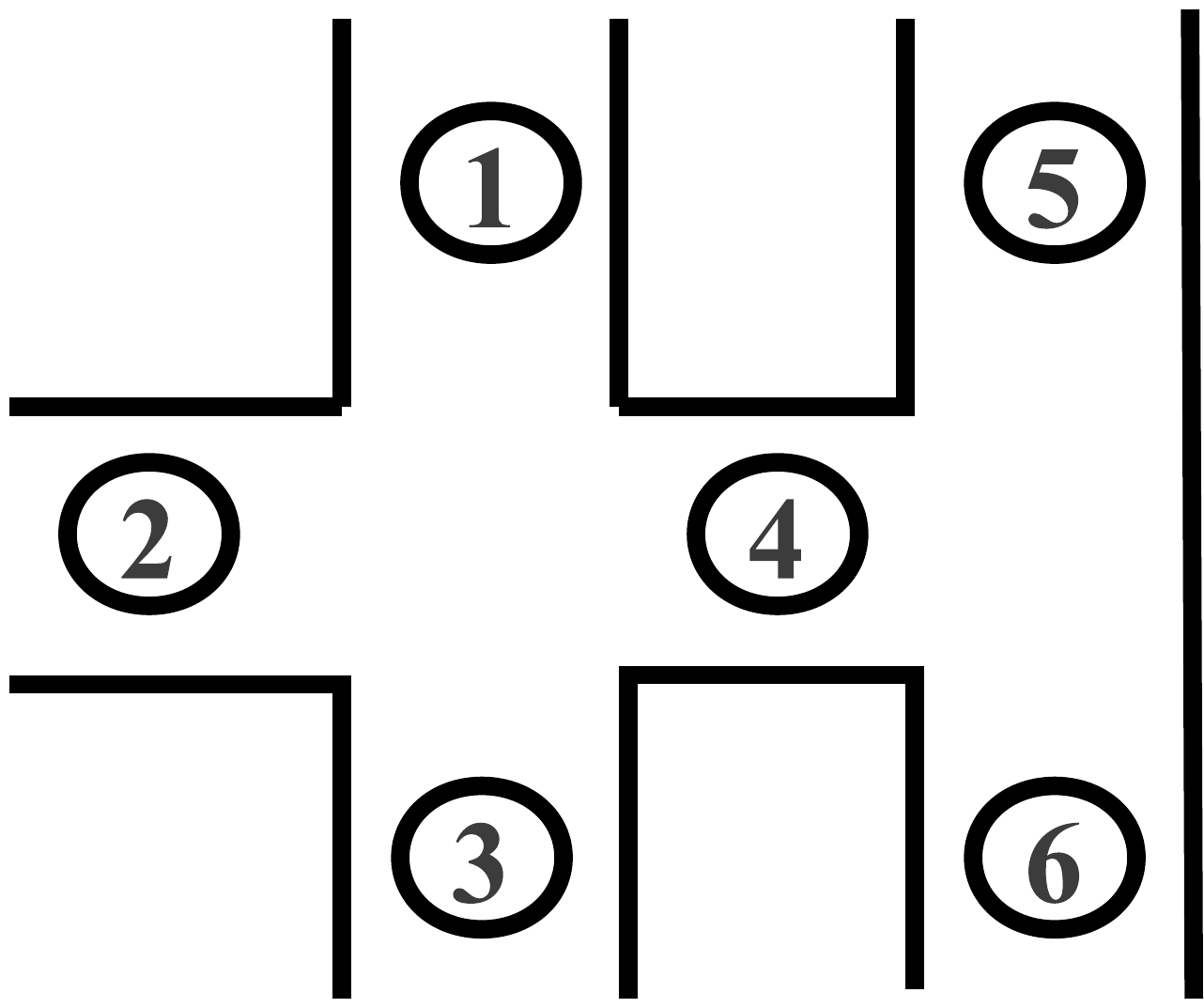}
      &
      \includegraphics[width=4cm]{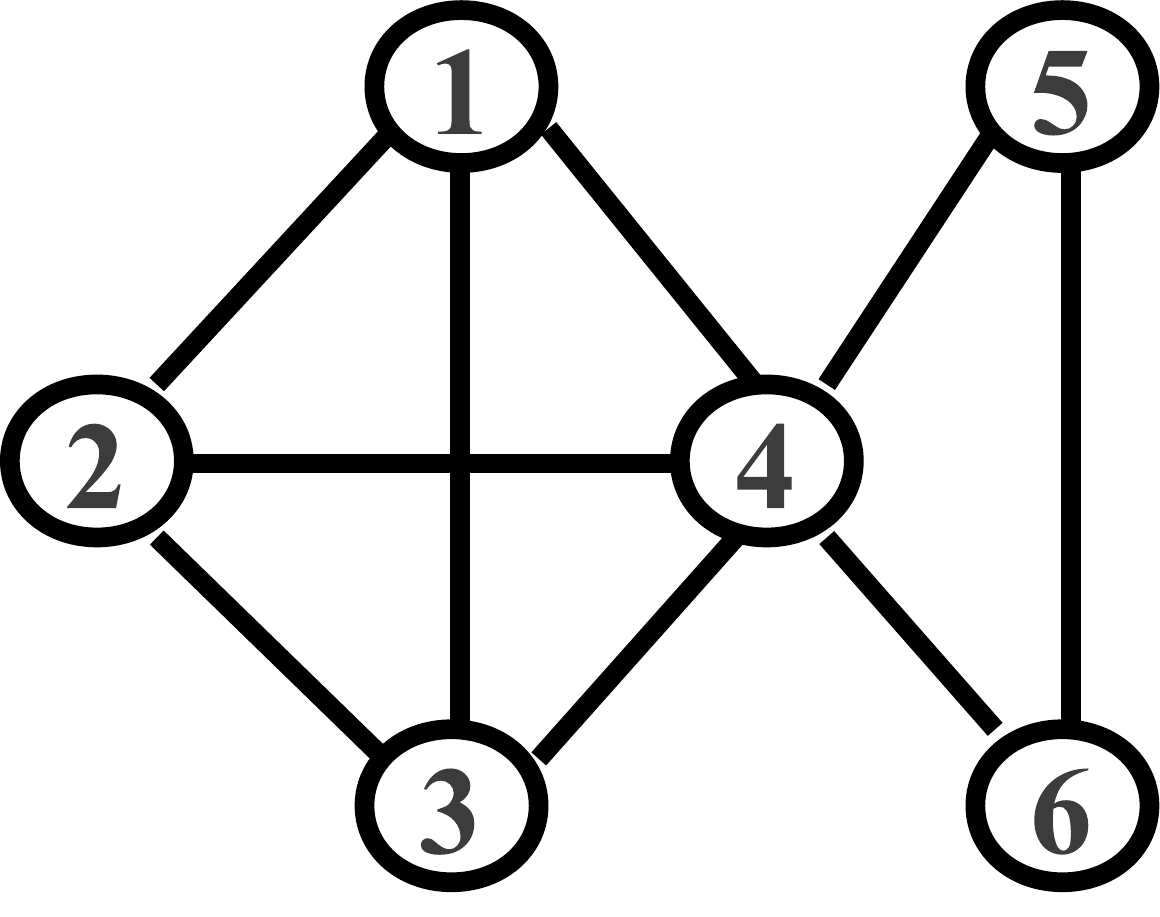}
    \end{tabular}
    \caption{A simple case to explain our graph representation of a road network. (a) A toy road network with six roads. (b) A graph representation of the toy road network consists of six vertices and nine edges.}\label{simple_net}
  \end{center}
\end{figure}

To explain our model, a simple case is shown in figure \ref{simple_net}.
There are six roads, represented as encircled numbers, and two intersections in figure \ref{simple_net} (a).
In this toy road network, vehicles on road 1 can directly move to roads 2, 3, or 4, but cannot move to road 5 and 6 without passing along road 4.
Then, this road network is translated to its graph representation, shown in figure \ref{simple_net} (b).
In this case, the joint probability density function is expressed as
\begin{eqnarray}
  P_{\mathrm{ex}}( \bm{x} ) := \frac{1}{Z_{\mathrm{ex}}} \prod_{i = 1}^6 \psi_i ( x_i ) \prod_{(i,j) \in E_{\mathrm{ex}}} \psi_{ij} ( x_i, x_j ) \label{prob_toy},\\
  Z_{\mathrm{ex}} := \int d \bm{x} \prod_{i = 1}^6 \psi_i ( x_i ) \prod_{(i,j) \in E_{\mathrm{ex}}} \psi_{ij} ( x_i, x_j )
\end{eqnarray}
where $E_{\mathrm{ex}} := \{ (1,2), (1,3), (1,4), (2,3), (2,4), (3,4), (4,5), (4,6), (5,6) \}$.
It should be noted that we ignore road direction relationships throughout this paper for simplicity, as shown in figure \ref{simple_net}; however extending to the model to include road direction relationships is straightforward.

\section{Traffic data reconstruction algorithm based on MRF} \label{reconstruction}
As mentioned in the introduction, a problem that affects traffic prediction is that we cannot collect complete traffic data of all roads , due to a lack of sensors.
Here after, $x_i$ is the traffic density of road $i$, which is obtained by dividing the number of vehicles by the length of road $i$.
In this section, we propose a method to reconstruct the traffic densities of unobserved roads from observed traffic densities based on MRF modeling and the Bayesian point of view.
Suppose that $\{ y^o_i | i \in V_o \subset V \}$ is a set of traffic densities of observed roads collected by sensors at a certain time and that we do not have complete information about all traffic densities.
Our goal is to reconstruct the traffic densities of unobserved roads $i \in V_u := V \backslash V_o $.

In the Bayesian point of view, a reconstruction of unobserved roads is inferred by using the posterior probability density function $P(\bm{x} \mid \bm{y} )$ expressed as
\begin{eqnarray}
  P(\bm{x} \mid \bm{y} ) := \frac{ P(\bm{y} \mid \bm{x} ) P(\bm{x}) }{ \int d\bm{x} P(\bm{y} \mid \bm{x} ) P(\bm{x}) }, \label{bayes}
\end{eqnarray}
where  $\bm{y} := \{ y_i \mid i \in V \} $ is the observed results for all roads collected by sensors, and $P(\bm{y} \mid \bm{x})$ is a conditional density function expressing how $\bm{y}$ is obtained from the true traffic density $\bm{x}$.
It should be noted that, since $\bm{y}$ is a specific value, a denominator in equation (\ref{bayes}) gives a constant value. 

To define a concrete joint probability density function of $\bm{x}$, we assume that the potential functions in equation (\ref{prob_model}) are expressed as
\begin{eqnarray}
\psi_i (x_i) := \exp \left( -\frac{\epsilon \eta}{2} x_i^2 +  \beta_i x_i \right) \label{pote1}
\end{eqnarray}
and
\begin{eqnarray}
\psi_{ij} ( x_i, x_j ) := \exp \left\{ -\frac{\eta}{2} \left( x_i - x_j \right)^2 \right\}, \label{pote2}
\end{eqnarray}
respectively, where $\{ \beta_i \in ( -\infty, \infty ) \mid i \in V \}$ and $\eta \in ( 0, \infty )$ are hyperparameters that determine features of road traffic.
$\beta_i$ represents how large a value the density of road $i$ takes and $\eta$ is associated with the closeness of neighboring roads in graph representation.

Then, the joint probability density function, which is regarded as the prior density function in the Bayesian point of view, of $\bm{x}$ is written as
\begin{eqnarray}
  P( \bm{x}; \bm{\beta}, \eta )
  &= \frac{1}{Z} \exp \left[ \sum_{i \in V} \beta_i x_i - \frac{\eta \epsilon}{2} \sum_{i \in V} x_i^2 - \frac{\eta}{2} \sum_{ (i,j) \in E } \left( x_i - x_j \right)^2 \right] \nonumber \\
  &= \sqrt{ \frac{ \eta^N \det C }{ \left( 2 \pi \right)^N } } \exp \left[ -\frac{ \eta }{ 2 } \left( \bm{x} - \frac{1}{\eta} C^{-1} \bm{\beta} \right)^{\mathrm{T}} C \left( \bm{x} - \frac{1}{\eta} C^{-1} \bm{\beta} \right) \right] \label{prior}
\end{eqnarray}
where $\bm{\beta} := \{ \beta_i | i \in V \}$ and the $N \times N$ matrix $C$ is defined by
\begin{eqnarray}
  C_{ij} := \cases{\epsilon + |\partial i|, &$i = j$\\-1, &$(i,j) \in E \ \mbox{or} \ (j,i) \in E$\\0, &$\mbox{otherwise}$} \label{mat_c}
\end{eqnarray}
where $\partial i := \{ j \in V | (i,j) \in E \ \mbox{or} \ (j,i) \in E \}$ is a set of vertices neighboring vertex $i$.
For positive $\epsilon$, the second term in the exponential in equation (\ref{prior}) guarantees the normalization of the joint probability density function.
This form of probability density function is known as a Gaussian MRF and has been widely used in various applications \cite{Rue_Held_Book_2005}.

We define a conditional density function $P(\bm{y} \mid \bm{x})$ as
\begin{eqnarray}
  P \left( \bm{y} \mid \bm{x} \right) &:= \prod_{i \in V} P \left( y_i \mid x_i \right), \label{likelihood} \\
  P \left( y_i \mid x_i \right) &\propto \cases{ 1, &$i \in V_u$ \\ \delta \left( y^o_i - x_i \right), &$i \in V_o$}
\end{eqnarray}
where $\delta \left( p - q \right)$ is the Dirac delta function.
Here, we assume that the unobserved traffic densities can take any real value with equal probability, and the densities of observed roads are not changed at all.

From equation (\ref{bayes}), the posterior probability density function $P \left( \bm{x} \mid \bm{y} \right)$ is written as $P \left( \bm{x} \mid \bm{y} \right) \propto P \left( \bm{y} \mid \bm{x} \right) P \left( \bm{x} \right)$.
Thus, the marginal posterior probability density function over the traffic densities of unobserved roads is expressed as
\begin{eqnarray}
  &P \left( \bm{x}_u \mid \bm{y}; \bm{\beta}, \eta \right) \nonumber \\
  &\propto \int d \bm{x}_o P \left( \bm{y} \mid \bm{x} \right) P \left( \bm{x}; \bm{\beta}, \eta \right) \nonumber \\
  &\propto \exp \left[ \sum_{i \in V_u} \beta_i x_i - \frac{\eta \epsilon}{2} \sum_{i \in V_u} x_i^2 - \frac{\eta}{2} \sum_{(i,j) \in E_1} \left( x_i -x_j \right)^2 - \frac{\eta}{2} \sum_{(i,j) \in E_2} \left( x_i - y^o_j \right)^2 \right] \nonumber \\
  &\propto \exp \left[ -\frac{\eta}{2} \left( \bm{x}_u - \frac{1}{\eta} A^{-1} \bm{b} \right)^{\mathrm{T}} A \left( \bm{x}_u - \frac{1}{\eta} A^{-1} \bm{b} \right) \right] \label{posterior}
\end{eqnarray}
where $\bm{x}_o := \{ x_i \mid i \in V_o \}$, $\bm{x}_u := \{ x_i \mid i \in V_u \}$, $E_1 := \{ (i,j) \in E \ \mbox{or} \ (j,i) \in E \mid i \in V_u, j \in V_u \}$ and $E_2 := \{ (i,j) \in E \ \mbox{or} \ (j,i) \in E \mid i \in V_u, j \in V_o \}$.
The $|V_u| \times |V_u|$ matrix $A$ and vector $\bm{b} := \{ b_i \mid i \in V_u \}$ are defined as follows:
\begin{eqnarray}
  A_{ij} &:= \cases{\epsilon + | \partial i |, &$i=j$ \\-1, &$(i,j) \in E_1 \ \mbox{or} \ (j,i) \in E_1$ \\0, &$\mbox{otherwise}$}, \label{mat_A} \\
  b_i &:= \beta_i + \eta \sum_{j \in \partial E_i } y^o_j 
\end{eqnarray}
where $\partial E_i := \{ j \in \partial i \mid (i,j) \in E_2 \ \mbox{or} \ (j,i) \in E_2 \}$.
The reconstruction of unobserved traffic densities in the RDB can be achieved to find values $\bm{x}_u^*$ such that
\begin{eqnarray}
  x_i^* := \cases{ x_i^\prime, &$x_i^\prime \geq 0$ \\ 0, &$x_i^\prime < 0$}, \label{esti_value}\\ 
  \bm{x_u}^\prime := \arg \max_{ \bm{x}_u } P \left( \bm{x}_u \mid \bm{y}; \bm{\beta}, \eta \right) \label{sub_esti_value}
\end{eqnarray}
for $i \in V_u$.
Because the marginal posterior probability density function in equation (\ref{posterior}) is a multivariate Gaussian distribution, values $\bm{x}_u^\prime$ are given by the mean vector of 
\begin{eqnarray}
  \bm{x}_u^\prime = \frac{1}{\eta} A^{-1} \bm{b}, \label{esti_v}
\end{eqnarray}
and we can calculate $\bm{x}_u^\prime$ exactly by applying the mean-field approximation \cite{Wainwright_Jordan_FDML_2008}.
The problem of estimating road traffic densities is reduced to solving the following simultaneous equations by an iteration method:
\begin{eqnarray}
  x_i^\prime = \frac{1}{\eta A_{ii} } \left( \beta_i + \eta \sum_{j \in \partial i } z_j \right) \label{iter}
\end{eqnarray}
for $i \in V_u$ where
\begin{eqnarray}
  z_i = \cases{ x_i^\prime, &$i \in V_u$ \\ y_i^o, &$i \in V_o$ }.
\end{eqnarray}
Since matrix $A$ is a sparse matrix, solving equation (\ref{iter}) is more efficient than calculating equation (\ref{esti_v}) directly.

The proposed algorithm for reconstructing the traffic densities of unobserved roads in an RDB is summarized as follows:
\begin{itemize} \setlength{\leftskip}{1cm}
  \item[{\bf Step 1.}] Determine the sets $V_o$ and $V_u$ from a graph representation of a road network. Input the values of observed traffic densities $\bm{y^o}$.
  \item[{\bf Step 2.}] Calculate matrix $A$ according to the definition equation (\ref{mat_A}).
  \item[{\bf Step 3.}] Solve the simultaneous equations in equation (\ref{iter}) by an iteration method, and then, use equation (\ref{esti_value}) and equation (\ref{sub_esti_value}) to obtain reconstructed traffic densities $\bm{x_u}^*$.
\end{itemize}

\section{Determining hyperparameters from HDB} \label{learning}
We derived a reconstruction algorithm for traffic densities of unobserved roads based on belief propagation described in the previous section.
However, we have not yet specified the values of the hyperparameters.
The purpose of this section is to show how we determine these parameters from the HDB using a machine learning method.
In this section, we assume that a large number of complete traffic data are available.
An explanation that excuses this assumption is that we do not need real complete data but artificial data to determine hyperparameters if it expresses the situations of road traffic well.  
And once we permit an assumption that daytime road traffic situations are similar on different days, we can create such pseudo complete traffic data at a certain time by merging the data collected on days because, different from the RDB, the HDB consists of many traffic data for long time periods and a comprehensive area. 
This assumption seems likely, especially at rush hour in an urban area where traffic predictions are necessary.
The extension to the area where this assumption is violated is mentioned in section \ref{disccusion} with its difficulty.

Let us suppose that we have $K$ complete road data of traffic densities, $\bm{d}^{(k)} := \{ d_i^{(k)} \in \left( -\infty, \infty \right) \mid i \in V \}, k = 1,\dots K$, created from the HDB.
The empirical distribution of the complete road data is given by
\begin{eqnarray}
  Q \left( \bm{x} \right) := \frac{1}{K} \sum_{k=1}^K \prod_{i \in V} \delta \left( x_i - d_i^{(k)} \right). \label{empiri}
\end{eqnarray}
A standard approach to determining hyperparameters is finding the one that maximizes the
likelihood function defined as 
\begin{eqnarray}
  L(\bm{\beta},\eta) := \int d \bm{x} Q( \bm{x} ) \log P ( \bm{x} ; \bm{\beta}, \eta ). \label{likefun}
\end{eqnarray}
However, this approach often give rise to the over-fitting problem, which occurs when the number of hyperparameters is larger than the number of data.
In the present model, there exits $N + 1$ hyperparameters.
Therefore, in the machine learning approach, we sometimes maximize the regularized likelihood function written as
\begin{eqnarray}
  L_\lambda (\bm{\beta},\eta ;\lambda ) := L(\bm{\beta},\eta) - \frac{\lambda}{2} \left( \eta^2 + \sum_{i \in V} \beta_i^2 \right). \label{rlike}
\end{eqnarray}
This regularization method is called ridge regression \cite{Hoerl_Kennard_Technometrics_1970}.
The parameter $\lambda$ is called the regularization parameter; it prevents the magnitudes of hyperparameters from being extremely large  to fit the data and is often determined by hand in advance.

From equation (\ref{prior}) and equation (\ref{likefun}), we can write equation (\ref{rlike}) as
\begin{eqnarray}
  L_\lambda (\bm{\beta},\eta ;\lambda ) =& - \sum_{i \in V} \beta_i \left< x_i \right>_D + \frac{ \eta }{ 2 } \sum_{i \in V} \left( \epsilon + | \partial i | \right) \left< x_i^2 \right>_D - \eta \sum_{ (i,j) \in E } \left< x_i x_j \right>_D \nonumber \\ &+ \frac{ 1 }{ 2 \eta } \bm{\beta}^\mathrm{ T } C^{-1} \bm{\beta} + \frac{ N }{ 2 } \log \eta + \frac{ \lambda }{ 2 } \sum_{ i \in V } \beta_i^2 + \frac{ \lambda }{ 2 } \eta^2 + \mbox{constant}
\end{eqnarray}
where the notation $\left< \cdots \right>_D$ denotes the expectation with respect to $Q \left( \bm{x} \right)$, i.e., the sample average of the complete traffic data set.
Using the gradient ascent method, we can obtain the values of $\bm{\beta}$ and $\eta$ that maximize $L_\lambda (\bm{\beta},\eta ;\lambda )$.
The gradient of $L_\lambda (\bm{\beta},\eta ;\lambda )$ with respect to $\bm{\beta}$ and $\eta$ are calculated as
\begin{eqnarray}
  \frac{ \partial L_\lambda (\bm{\beta},\eta ;\lambda ) }{ \partial \beta_i } =& - \left< x_i \right>_D + \frac{ 1 }{ \eta }\sum_{j \in V} C^{-1}_{ij} \beta_j + \lambda \beta_i, \label{gradi_beta} \\
  \frac{ \partial L_\lambda (\bm{\beta},\eta ;\lambda ) }{ \partial \eta } =& \frac{ 1 }{ 2 } \sum_{i \in V} \left( \epsilon + |\partial i| \right) \left< x_i^2 \right>_D - \sum_{ (i,j) \in E } \left< x_i x_j \right>_D \nonumber \\ &- \frac{ 1 }{ 2 \eta^2 } \bm{\beta}^\mathrm{ T } C^{-1} \bm{\beta} + \frac{ N }{ 2 \eta } + \lambda \eta. \label{gradi_eta}
\end{eqnarray}
It should be noted that, although we need the inverse of matrix $C$ in equation (\ref{gradi_beta}) and equation (\ref{gradi_eta}), it is enough to calculate the inverse matrix once in pre-processing because it depends on only the structure of a given road network.

\section{Numerical experiments} \label{num_exp}
\begin{figure}
  \begin{center}
    \includegraphics[width=9cm]{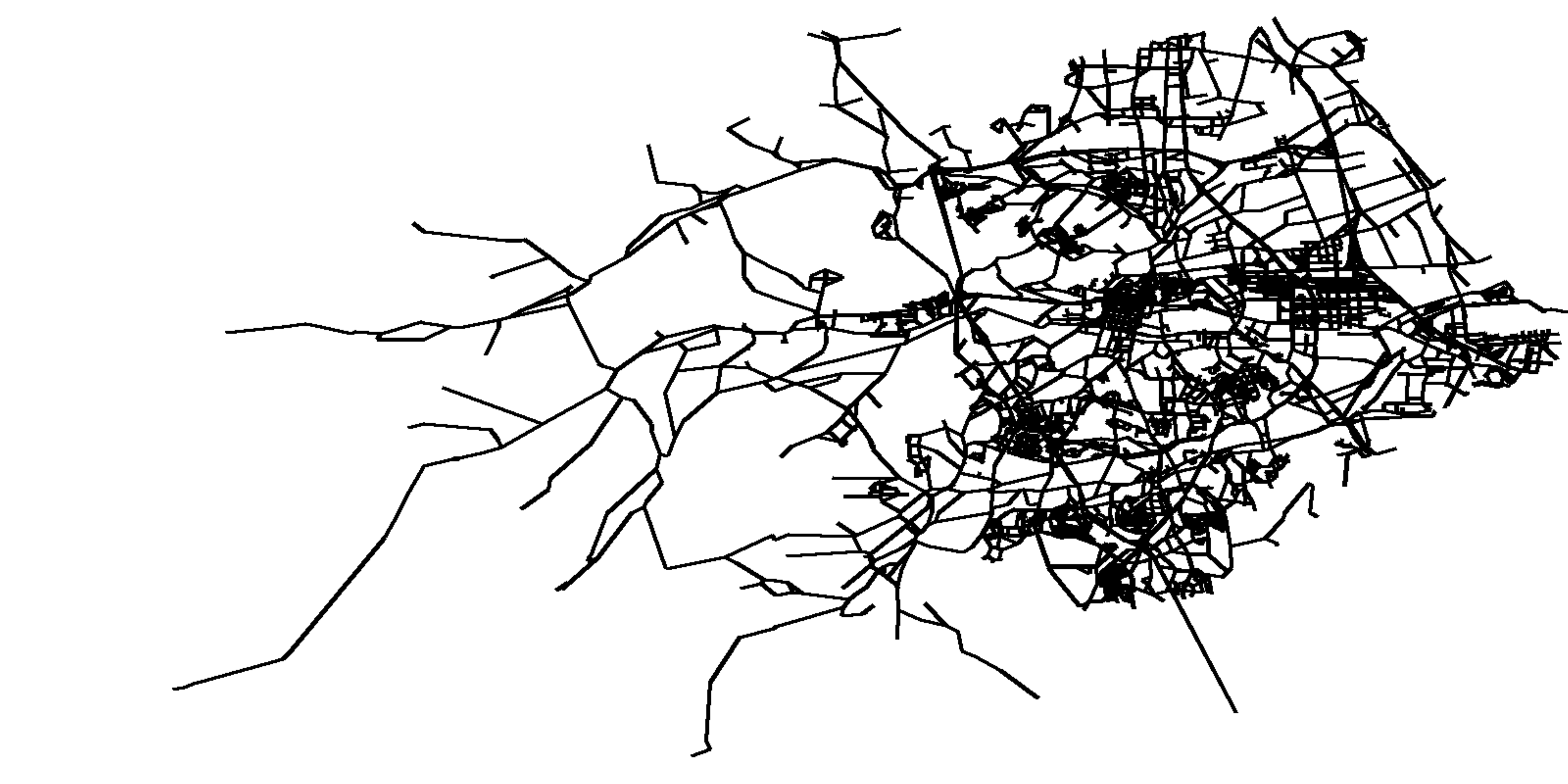}
    \caption{Road network of Sendai, Japan that we used in numerical experiments. There are $9582$ vertices and $20482$ edges in the graph representation of this road network.}\label{sendai}
  \end{center}
\end{figure}

In this section, we describe the numerical verification of the performance of our MRF model.
We used the real road network of Sendai, Japan, described in figure \ref{sendai}, and $360$ vehicle traffic data, which constitute a snapshot of its simulated vehicle traffic.
These simulation data represent the real vehicle traffics in Sendai, Japan.
In the graph representation of the Sendai road network, there are $9582$ vertices and $20482$ edges.

To evaluate the performance of our model, we conducted leave-one-out cross-validation \cite{Bishop_Book_2006} in which only one data item is used to check the performance, and the others are used to determine the hyperparameters.
The performance of the model is then given by the average over all the choices of test data.
That is, in each choice of test data, we regard the other data as complete data created from the HDB, and the test data are used to create the data in the RDB.
In the test phase, we randomly selected unobserved roads with equal probability $p$ from all roads, and then, reconstructed the traffic densities of the unobserved roads using our algorithm.
In each test data, we evaluated the performance of our model by the average of mean absolute errors (MAE) between the true and reconstructed traffic density over $500$ trials defined by
\begin{eqnarray}
  \mbox{[MAE]}_m := \frac{1}{500} \sum_{l=1}^{500} \left( \frac{1}{N_l} \sum_{i=1}^{|V|} \left| x_i^* - x_i^{(m)} \right| \right)
\end{eqnarray}
where $N_t$ is the number of unobserved roads at the $l$th trial and $x_i^{(m)}$ is the true traffic density of road $i$ in the $m$th data.
Hence, the results of leave-one-out cross-validation are given by
\begin{eqnarray}
  \mbox{MAE} := \frac{1}{360} \sum_{m=1}^{360} \mbox{[MAE]}_m
\end{eqnarray}
for each $\lambda$.

\begin{figure}
  \begin{center}
    \includegraphics[width=8cm]{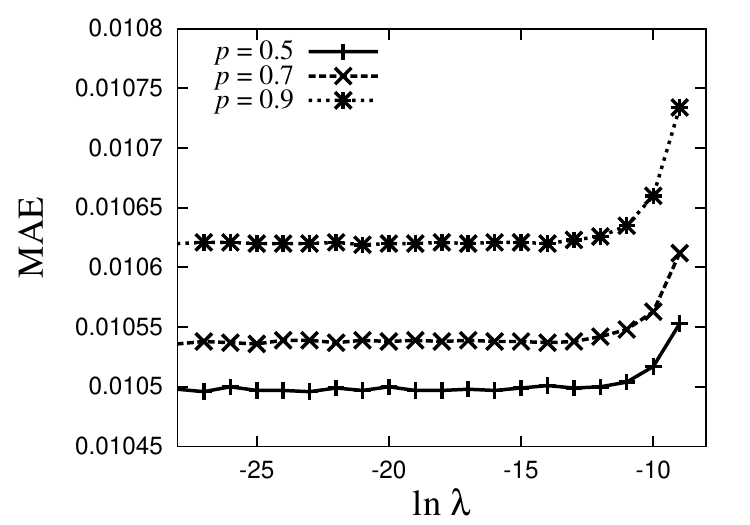}
    \caption{MAE versus $\ln \lambda$ when $p = 0.5$, $p = 0.7$, and $p = 0.9$. Each point is obtained by averaging over 500 trials.}\label{various_lambda}
  \end{center}
\end{figure}

Figure \ref{various_lambda} shows the plot of MAE versus $\ln \lambda$ when $p = 0.5$, $p = 0.7$, and $p = 0.9$.
Here, we set $\epsilon = 10^{-4}$ in equation (\ref{pote1}) so that the effect of the first term in the exponent is as small as possible, because this term is needed only to guarantee the normalization.  
In the region where $\ln \lambda$ is sufficiently small, our reconstruction algorithm yields a good performance for all values of $p$, and the MAE approach asymptotically to the values when $\lambda = 0$.
When $\lambda = 0$, MAE was $0.01049$, $0.01053$, and $0.01062$ for $p = 0.5$, $p = 0.7$, and $p = 0.9$, respectively.
Hence, $\lambda = 0$ is the best approach for determining the hyperparameters in our model.

\begin{figure}
  \begin{center}
    \begin{tabular}{ c c }
      (a) & (b)\\
      \includegraphics[width=6cm]{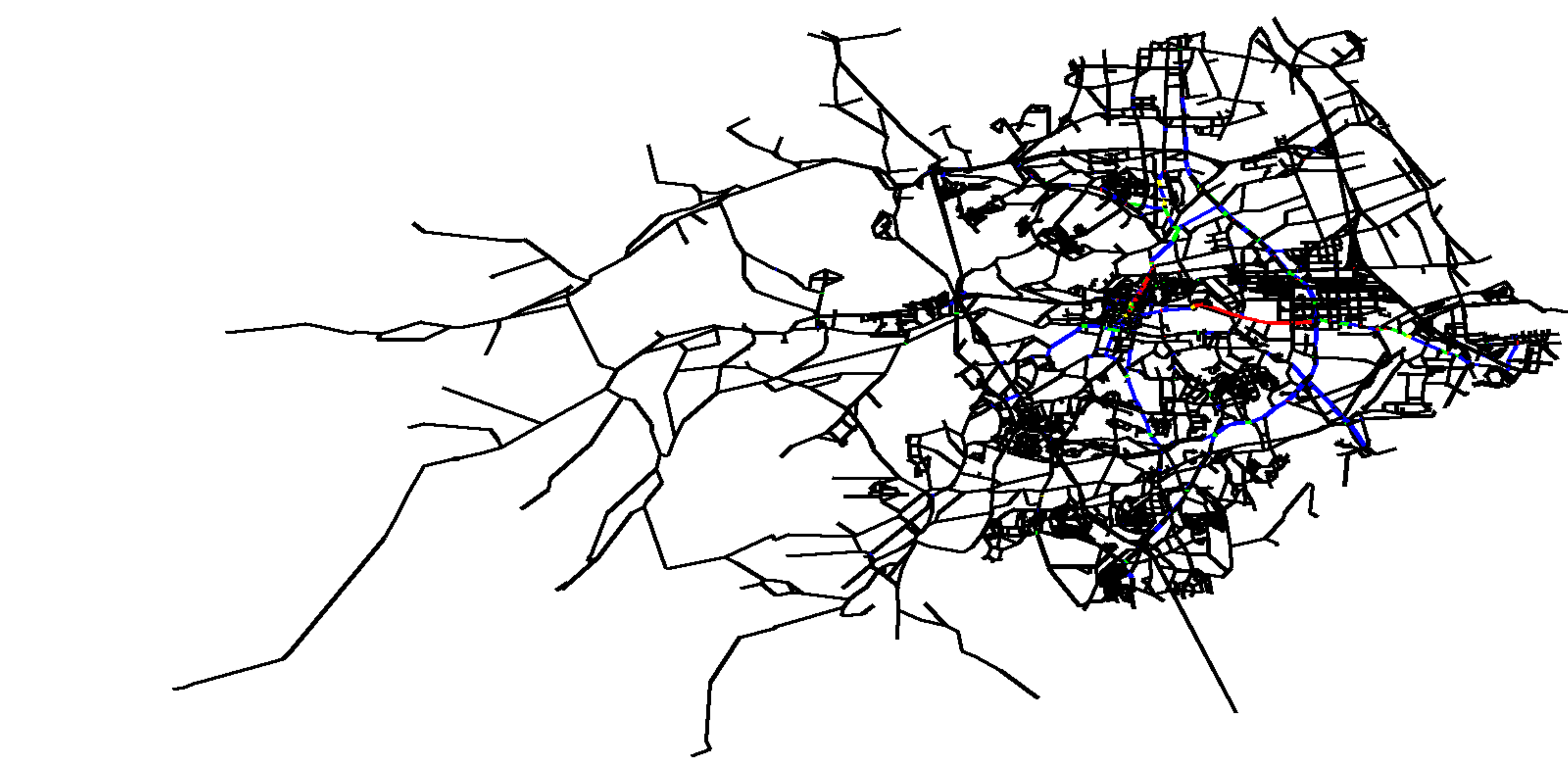}
      &
      \includegraphics[width=6cm]{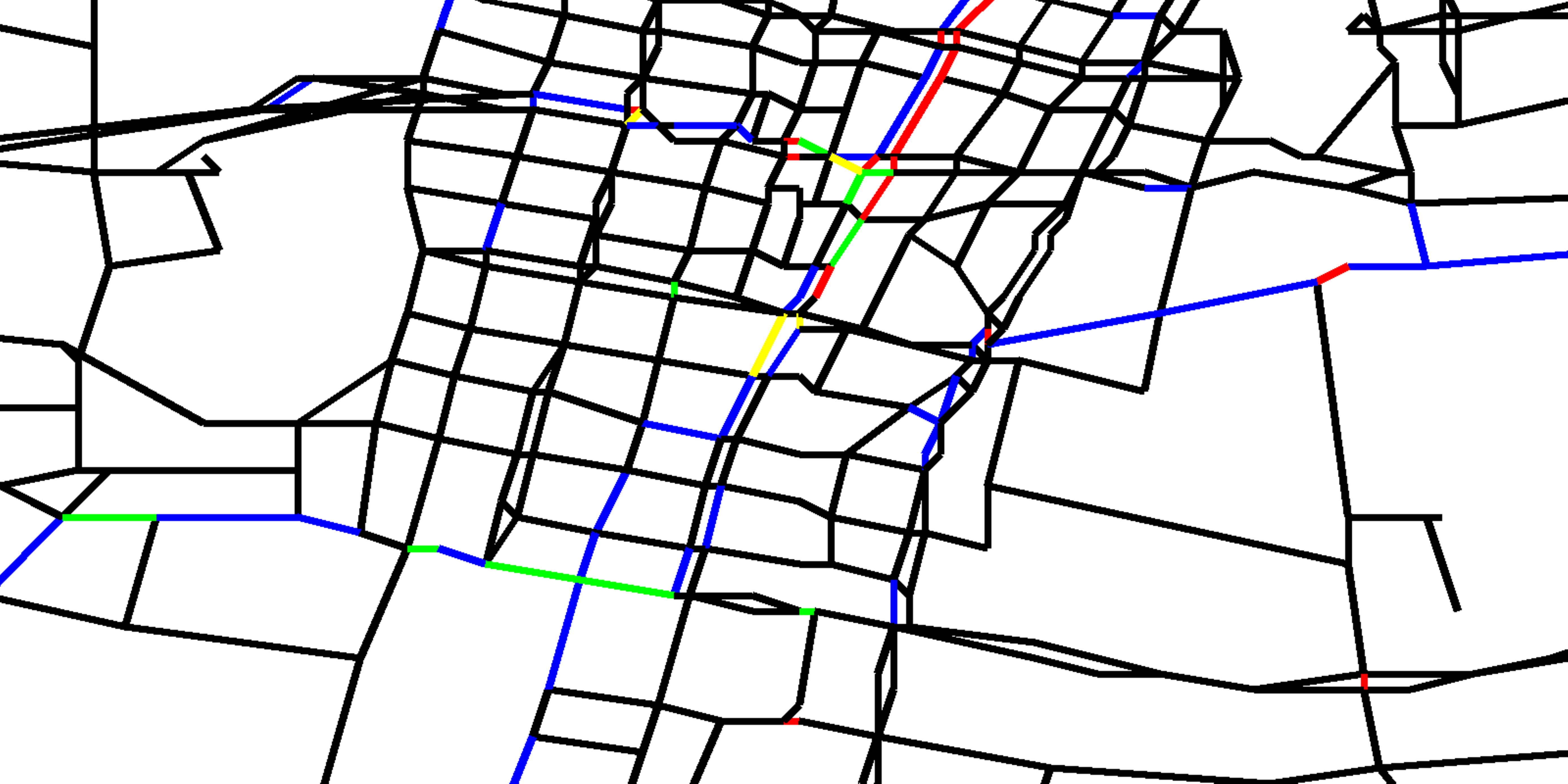}\\
      (c) & (d)\\
      \includegraphics[width=6cm]{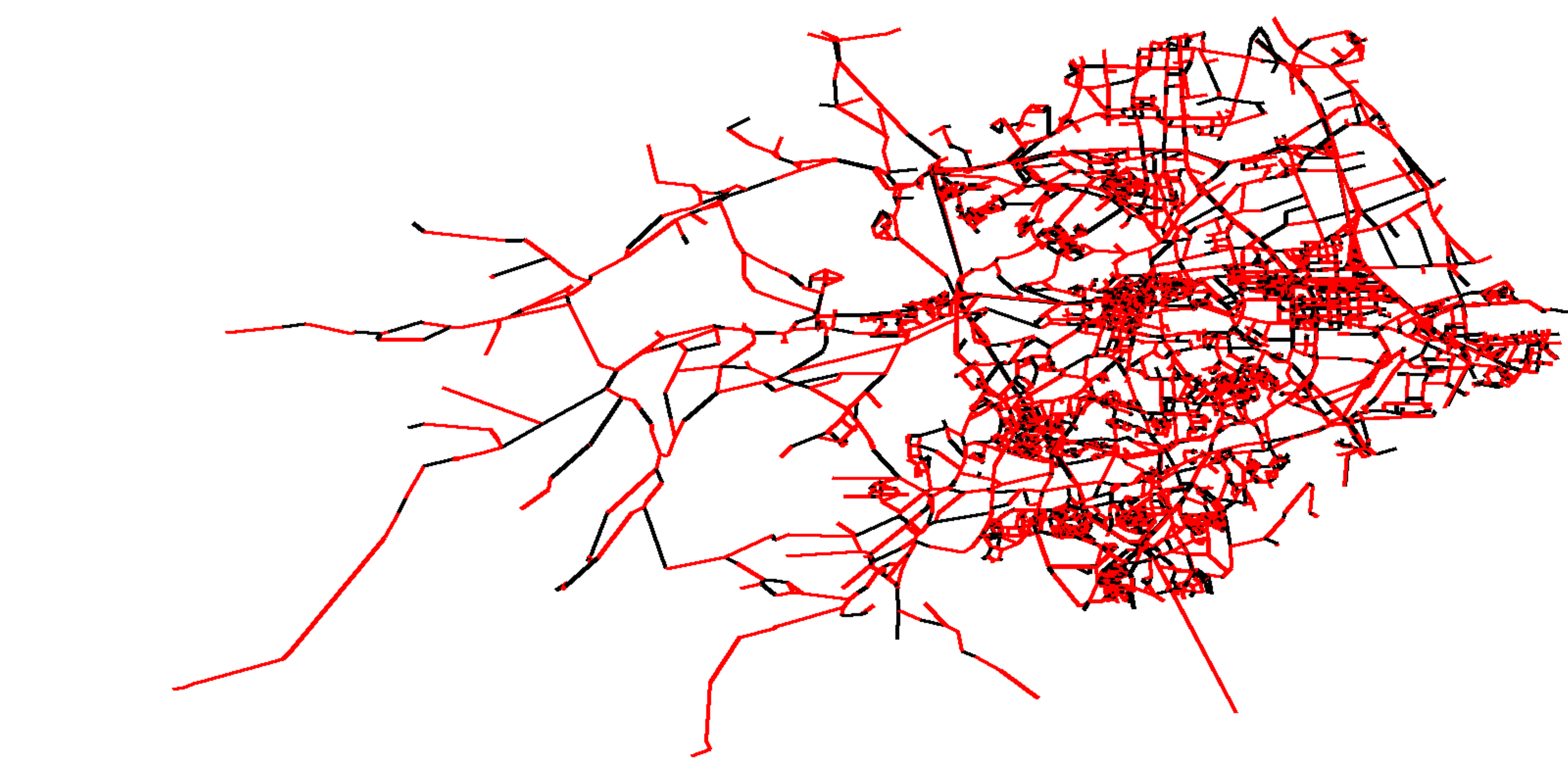}
      &
      \includegraphics[width=6cm]{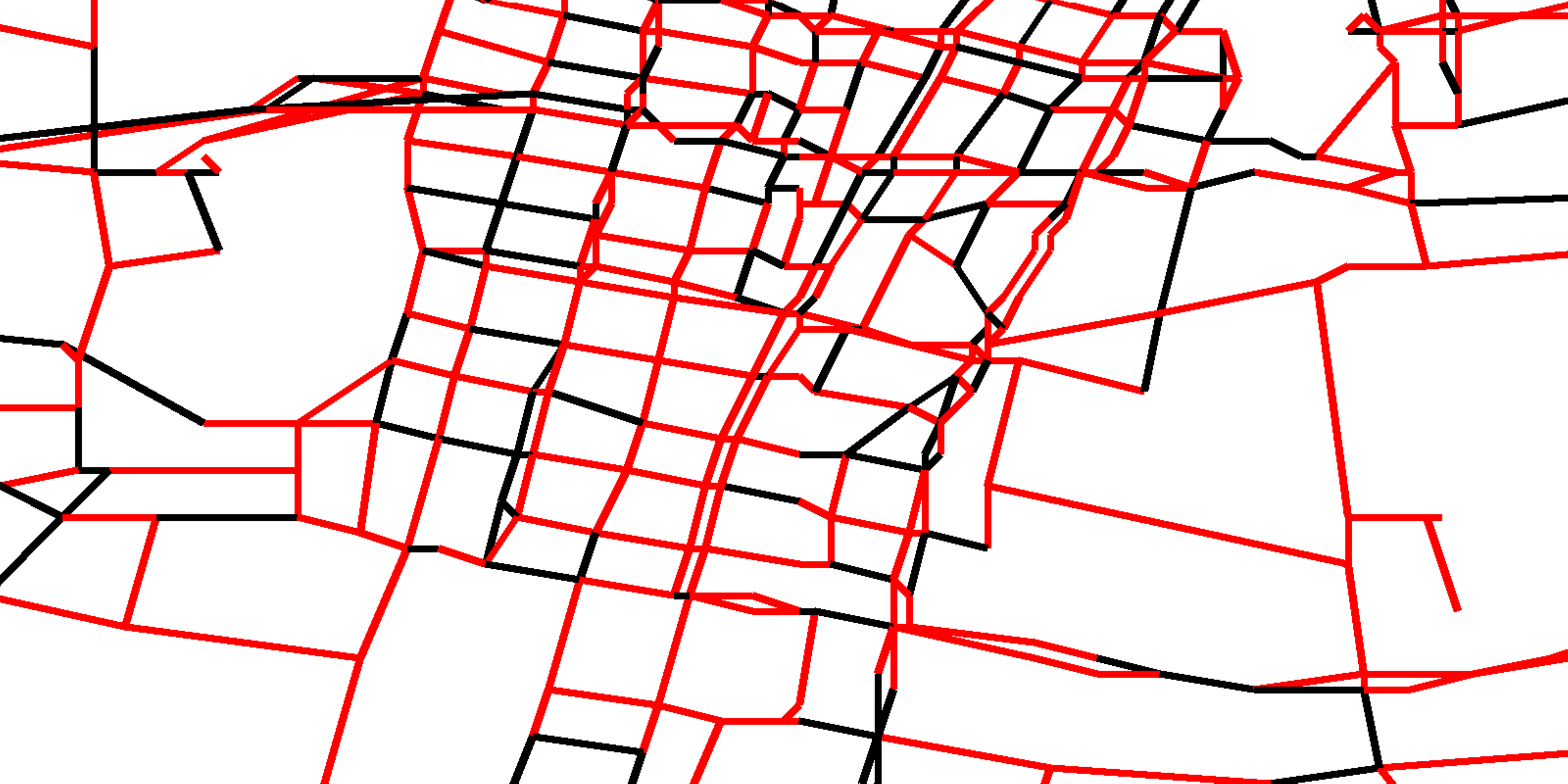}\\
      (e) & (f)\\
      \includegraphics[width=6cm]{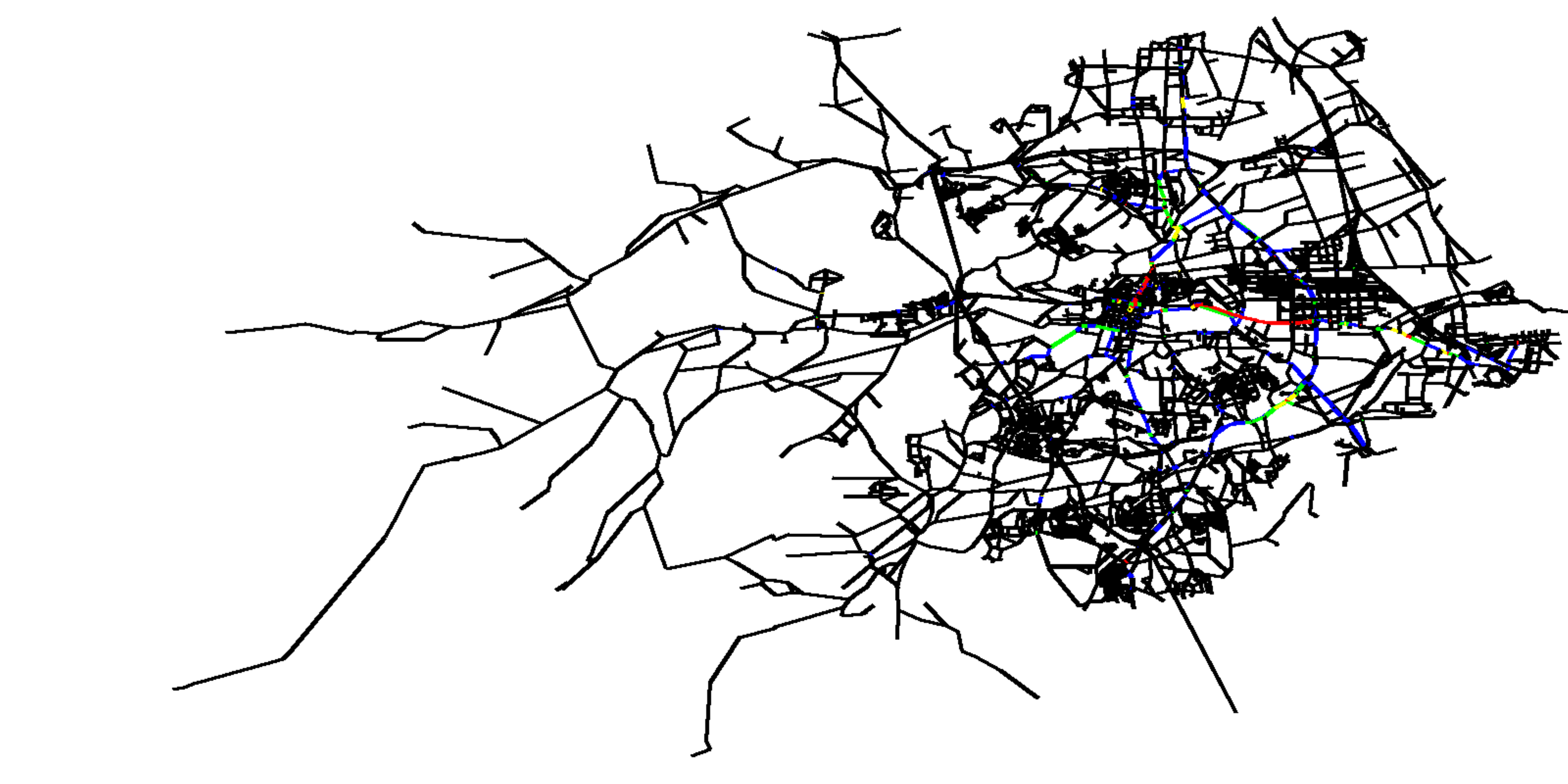}
      &
      \includegraphics[width=6cm]{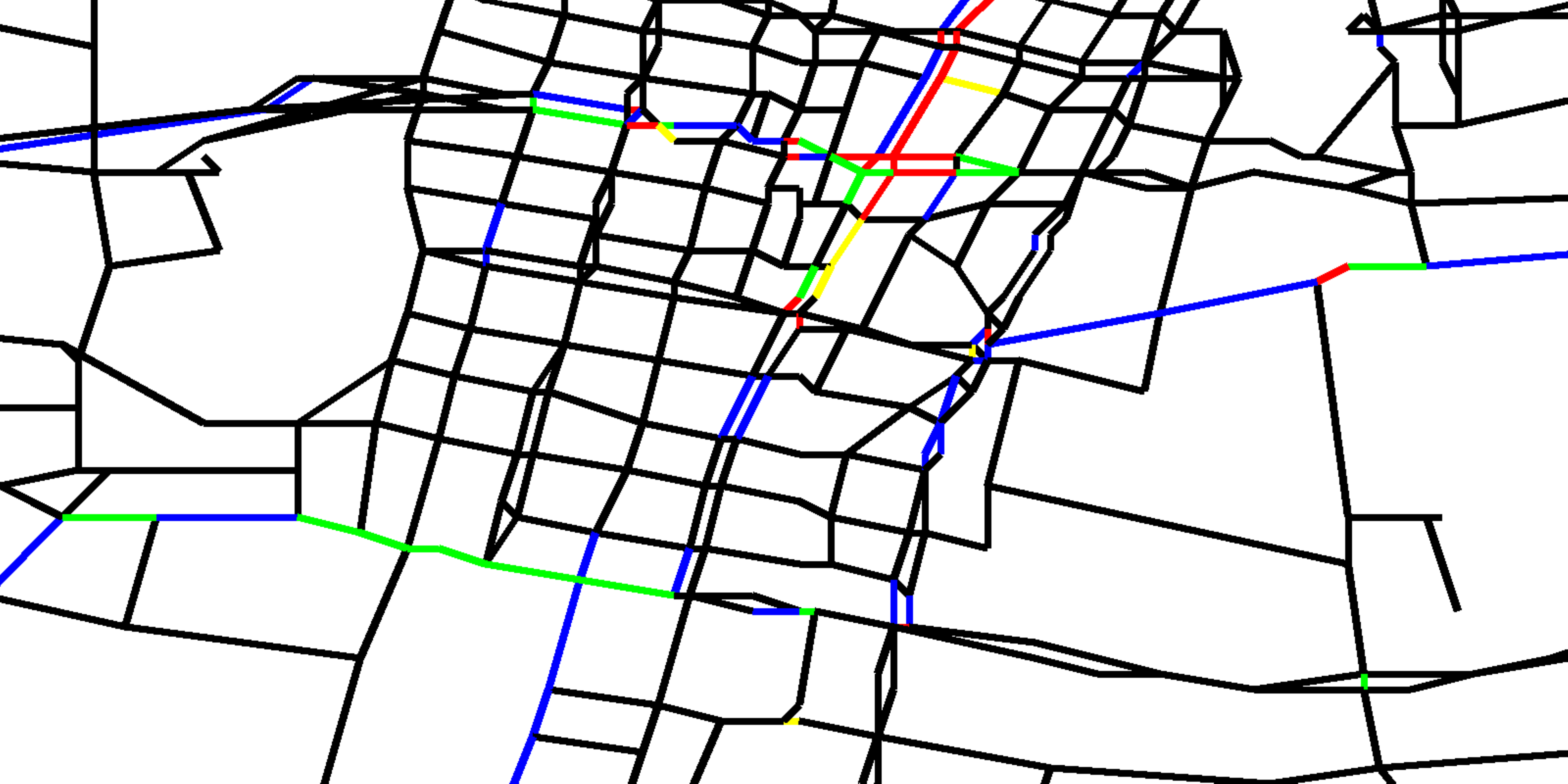}\\
    \end{tabular}
    \includegraphics[width=7cm]{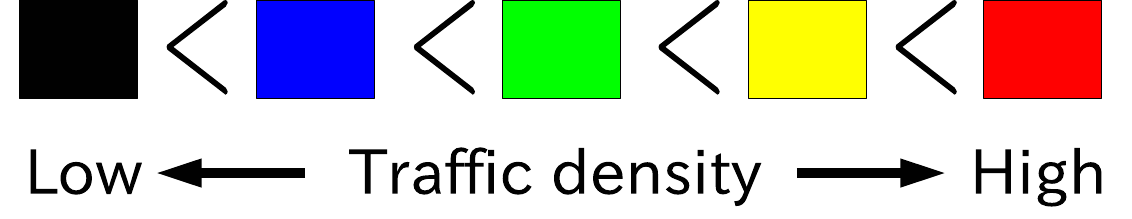}
    \caption{An example of our numerical experiments using simulated data for the road network of Sendai, Japan. (a) True traffic density data where each road is colored according to its traffic density. (b) Enlarged image of a part of (a). (c) Positions of unobserved roads where unobserved roads are colored red. About $70\%$ of roads in this road network are unobserved. (d) Enlarged image of part of (c). (e) Result of reconstruction using our model. The MAE between (a) and these result is $0.0106$. (f) Enlarged image of part of (e).}\label{demo}
  \end{center}
\end{figure}

Figures \ref{demo} shows an example of our numerical experiments when $p = 0.7$ and $\lambda = 0$.
Figure \ref{demo} (a) shows the original traffic densities and figure \ref{demo} (e) shows the reconstructed traffic densities using our model.
In figure \ref{demo} (a) and (e), the road colors are changed from black to blue, green, yellow, and red in order of increasing traffic densities by $0.05$ intervals where a black road is one where the traffic density takes a value between $0.0$ and $0.05$.
Figure \ref{demo} (c) shows the positions of unobserved roads; we colored the roads red when they were selected as unobserved roads with probability $p = 0.7$.
That is, about $70 \%$ of roads are unobserved.
The black roads in figure \ref{demo} (c) denote the positions of traffic sensors that collect traffic densities of the observed parts in the RDB.
The MAE between figure \ref{demo} (a) and (e) is $0.0106$.
Figure \ref{demo} (b), (d), and (f) are enlarged images of the downtown area of Sendai, Japan shown in figure \ref{demo} (a), (c) and (e), respectively.
Our method yields good reconstruction results of the unobserved parts in the RDB, as shown in figure \ref{demo}, and these results show that our Gaussian model well expresses the real traffic situation.

\section{Concluding Remarks} \label{disccusion}
In this paper, we proposed a traffic data reconstruction method based on MRF modeling.
The reconstruction of unobserved parts in an RDB is reduced to simple simultaneous equations of mean-field method.
The hyperparameters in our model are determined utilizing past traffic data in an HDB.
We checked the performance of our method by conducting leave-one-out cross-validation, as described in section \ref{num_exp}.
In the numerical experiments, we used large scale simulated data in Sendai, Japan.
We think it difficult to apply previous reconstruction method to such a large scale road network.
It should be noted that, in this study, we reconstructed only the traffic density data, but the extension of our MRF model to other data types, such as speed or flow, and furthermore, to combinations of these data types is straightforward.

In our scheme, we made two assumptions about the HDB and traffic densities for analytical convenience.
The first assumption was that we can create a number of complete traffic data from an HDB because it can contain many traffic data for a long time period and comprehensive area, and the daily conditions of road traffic seem similar, especially in an urban area.
This assumption might be perfunctory in an area where the amount of traffic is small, as in a rural area.
We can modify our learning framework by using an expectation maximization algorithm \cite{Dempster_Laird_Rubin_JRSS_1977} for determining hyperparameters from an incomplete data set in an HDB of such an area.
However, we need to calculate the inverse of $K$ different matrices $A_k \ (k = 1, \dots, K)$ in this framework. 
The matrix $A_k$ is defined similar as equation (\ref{mat_A}) but dimension corresponding to the number of unobserved roads in $k$th data may be different.
Therefore, analytical treatment is distant and we need to seek some approximate method to calculate $A_k^{-1}$ in this framework.
It should be noted that the reconstruction scheme described in section \ref{reconstruction} does not change after this modification.
The second assumption was that traffic density can take any real value, and its potential functions have quadratic form, as equations (\ref{pote1}) and (\ref{pote2}).
This assumption allows the Gaussian MRF of traffic densities, which is a single mode density function. 
In our definition of MRF modeling of traffic in section \ref{modeling}, we did not need to restrict the form of the potential functions and their arguments.
One extension that would result in a more complex MRF is using non-negative Boltzmann machine \cite{Downs_Mackay_Lee_AdvanceNIPS_2000} which is a multi-modal density function for the joint density function of $\bm{x}$; however we need an approximation method \cite{Yasuda_Tanaka_PhilosophicalMagazine_2012} because its analytical treatment is difficult. 
We aim to develop our MRF model further in these directions.

\section*{Acknowledgements}
The authors thank Prof. Masao Kuwahara and Dr. Jinyoung Kim of the Graduate School of Information Science, Tohoku University, for providing road network data and traffic simulation data.
This work was partly supported by Grants-In-Aid (Nos. 25280089, 24700220 and 25$\cdot$7259 ) from the Ministry of Education, Culture, Sports, Science and Technology of Japan.
S.K. was partially supported by a Research Fellowships of Japan Society for the Promotion of Science for Young Scientists.

\section*{References}
\bibliographystyle{iopart-num}
\bibliography{references}

\end{document}